\documentclass[10pt,twocolumn,letterpaper]{article}

\usepackage{cvpr}              
\usepackage{graphicx}
\usepackage{amsmath}
\usepackage{amssymb}
\usepackage{booktabs}
\usepackage{multirow}
\usepackage{makecell}
\usepackage{comment}
\usepackage[accsupp]{axessibility}
\usepackage[pagebackref,colorlinks]{hyperref}
\usepackage[capitalize]{cleveref}
\crefname{section}{Sec.}{Secs.}
\Crefname{section}{Section}{Sections}
\Crefname{table}{Table}{Tables}
\crefname{table}{Tab.}{Tabs.}

\def\cvprPaperID{5933} 
\def\confName{CVPR}
\def\confYear{2022}

\begin{document}

\title{Multi-View Transformer for 3D Visual Grounding}

\author{Shijia Huang \and Yilun Chen \and Jiaya Jia \and Liwei Wang \and
The Chinese University of Hong Kong \\
 {\tt\small \{sjhuang,ylchen,leojia,lwwang\}@cse.cuhk.edu.hk}
}

\maketitle

\begin{abstract}
The 3D visual grounding task aims to ground a natural language description to the targeted object in a 3D scene, which is usually represented in 3D point clouds. Previous works studied visual grounding under specific views. The vision-language correspondence learned by this way can easily fail once the view changes. In this paper, we propose a Multi-View Transformer (MVT) for 3D visual grounding. We project the 3D scene to a multi-view space, in which the position information of the 3D scene under different views are modeled simultaneously and aggregated together. The multi-view space enables the network to learn a more robust multi-modal representation for 3D visual grounding and eliminates the dependence on specific views. Extensive experiments show that our approach significantly outperforms all state-of-the-art methods. Specifically, on Nr3D and Sr3D datasets, our method outperforms the best competitor by $11.2\%$ and $7.1\%$ and even surpasses recent work with extra 2D assistance by $5.9\%$ and $6.6\%$. Our code is available at \href{https://github.com/sega-hsj/MVT-3DVG}{https://github.com/sega-hsj/MVT-3DVG}.
\end{abstract}

\section{Introduction}
\label{sec:intro}

Visual Grounding (VG) aims to ground a natural language description to the target object. The field has made tremendous progress in 2D images~\cite{plummer2015flickr30k,kazemzadeh2014referitgame,yu2016modeling,wang2021improving}. With the rapid development of 3D sensors and 3D vision technology, 3D visual grounding has recently attracted more attention, with wide applications including vision-language navigation~\cite{wang2019reinforced,zhu2020vision}, intelligent agents~\cite{savva2019habitat,xia2018gibson}, and autonomous vehicles~\cite{feng2021cityflow,mittal2020attngrounder}. Compared to 2D visual grounding, the 3D task has more complex input data (e.g., sparse point clouds) and more variant spatial relationships, switching the output from grounding to 2D regions to 3D objects.

Recent works~\cite{yuan2021instancerefer,achlioptas2020referit3d,chen2020scanrefer,he2021transrefer3d,roh2021languagerefer,yang2021sat} mainly follow a two-stage scheme, i.e., first generating all candidate objects in 3D scene and then selecting the most matched one. These two-stage approaches in 3D visual grounding are tailored from 2D visual grounding methods~\cite{yu2018mattnet,liu2019learning}, which have not considered the unique property of 3D data. Therefore, in this paper, we bring the community's attention back to studying the intrinsic property of 3D data to break through the 3D visual grounding research.

\begin{figure}[t]
  \centering
  \includegraphics[width=0.45\textwidth]{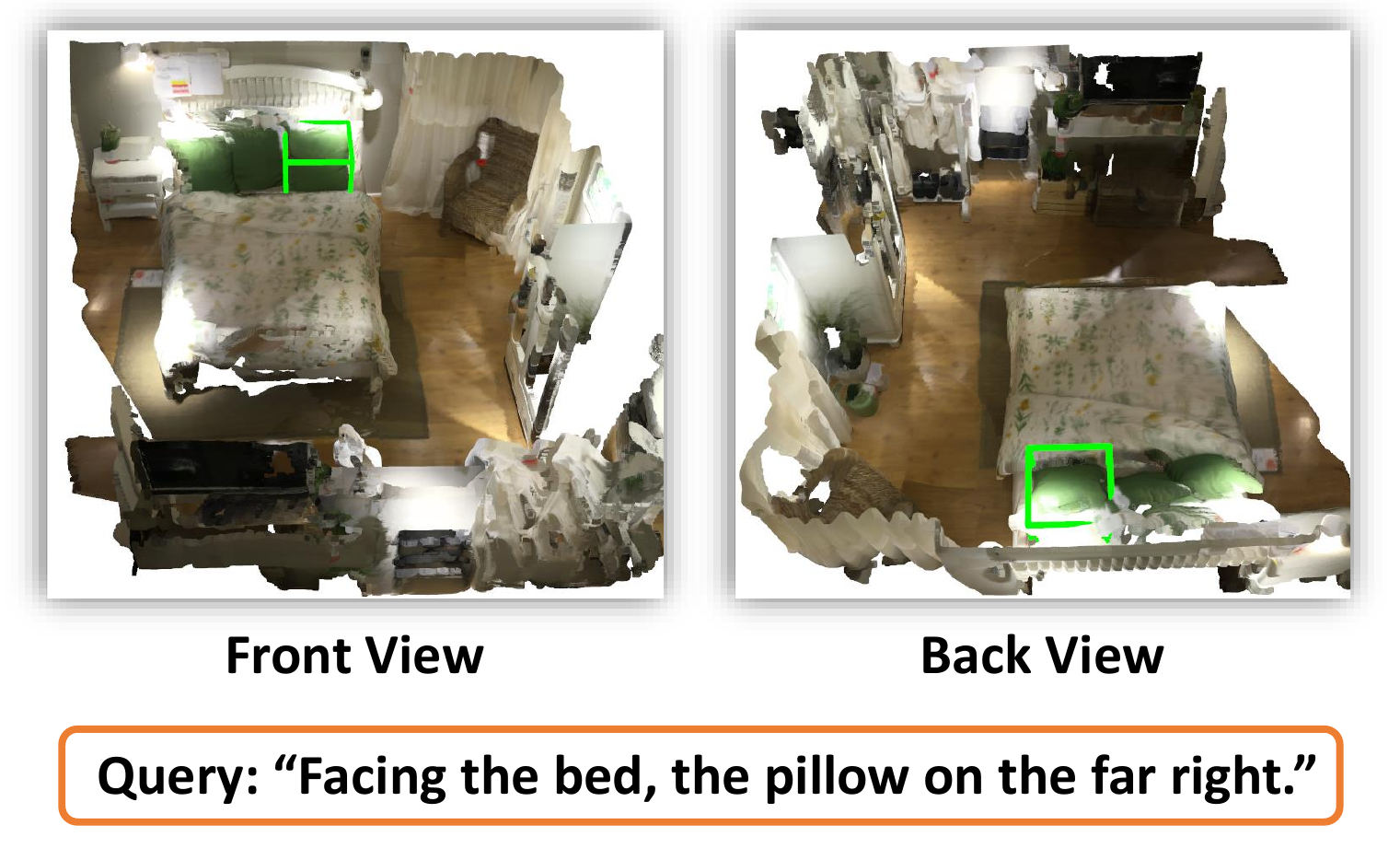}
  \vspace{-5pt}
  \caption{ An example from Nr3D~\cite{achlioptas2020referit3d}: when the view is changed, the position information in 3D scene can be quite different. The mentioned ``pillow'' exhibits changed positions, i.e., 3D coordinates, when the view is changed from the front to the back.
  }
  \label{fig:motivation}
  \vspace{-5pt}
\end{figure}

Position information is essential to locate an object. The visual grounding aligns visual position information in the scene (e.g., coordinates of objects) with the location described in natural language (e.g., ``left'' or ``right''). Unlike 2D images with static views, 3D scenes can freely rotate to different views. These dynamic view changes make 3D visual grounding more challenging. The view changes will directly affect the position encoding, e.g., coordinates of objects, thus bringing more difficulty in representing 3D objects. For example, as shown in Fig.~\ref{fig:motivation}, the rightmost pillow in the front view becomes the leftmost one in the back view. This discrepancy commonly exists in real applications: when a robot or an intelligent agent is navigating in the 3D scene, its view is usually different from the speaker's or commander's view. Such view discrepancy can bring more ambiguity in identifying the location of targets. To alleviate this view issue, when Referit3D~\cite{achlioptas2020referit3d} were collecting data, they split the data into view-dependent and view-independent categories according to whether the description was constrained to the speaker's view. They have provided additional view guidance, e.g., ``Facing the bed'' in the example of Fig.~\ref{fig:motivation}, to those view-dependent data. In practice, the view constraint is usually implicit, and we cannot always rely on speakers to provide augmented descriptions to solve the problem. Especially, the robot or the intelligent agent should be able to ground to the target object no matter how its view is different from the speaker's. Thus, developing a robust approach to view changes becomes imperative in 3D visual grounding.

Existing methods~\cite{achlioptas2020referit3d,roh2021languagerefer,he2021transrefer3d,yuan2021instancerefer,zhao20213dvg,yang2021sat,feng2021free} have paid little attention to the issue as mentioned above and mainly learned the grounding task under specific views. The vision-language correspondence learned in this way can easily fail once the view changes. An intuitive solution to alleviating the problem is to add augmented rotation data under different views during training. However, as we tried and tested (see Tab.~\ref{tab:component_ablation} in our experimental section), such a straightforward method does not bring the improvement. This result can be understood, i.e., although novel views of the 3D scene can be augmented in the training stage, visual features under specific views are not accurately aligned with the position information embedded in the language description. An alternative is to ask the model to predict the view attached to the language query, say, the speaker's view, and then align the query with the visual features under the predicted view. However, this will incur a lot more costs for view prediction, and if the model predicts the wrong view, the alignment between the two modalities will inevitably be damaged. Experiments provided by LanguageRefer~\cite{roh2021languagerefer} confirmed this point, i.e., even if they manually correct the view of all training data to the speaker's view, the performance of the model on unseen test data with view discrepancy still cannot be improved. Therefore, in this paper, we turn to the exploration of learning better multi-modal representations for 3D visual grounding, which are robust to view changes, instead of augmenting view data or predicting views.

In this paper, we propose a Multi-View Transformer (MVT) for 3D visual grounding to learn such a view-robust representation. Given the 3D scene under a specific view, our MVT projects it to a multi-view space by rotating the original view with equal angles. Our approach simultaneously models the position information within the 3D scene under different views and aggregates all information to learn a view-robust representation by eliminating the dependence on the starting view. Furthermore, to reduce the computational cost of modeling multi-view space, we decouple the 3D object representation computation into computing point clouds features and object coordinates separately to share the point clouds features across views. Besides, with the rich and fine-grained category labels provided by Nr3D~\cite{achlioptas2020referit3d} and Sr3D~\cite{achlioptas2020referit3d}, we make full use of these language priors and propose a language-guided object classification task to enhance the object encoder further. 
Extensive experiments have been conducted on three mainstream benchmark datasets Nr3D~\cite{achlioptas2020referit3d}, Sr3D~\cite{achlioptas2020referit3d}, and ScanRefer~\cite{chen2020scanrefer}. Our MVT outperforms all state-of-the-art methods by a large margin. Specifically, on Nr3D and Sr3D, MVT not only outperforms the best competitor by $11.2\%$ and $7.1\%$ respectively but also surpasses the method~\cite{yang2021sat} with extra 2D assistance by $5.9\%$ and $6.6\%$.


\section{Related Work}
\noindent\textbf{3D Visual Grounding.} The task of 3D visual grounding aims to ground objects described by natural language in 3D scenes. Referit3D~\cite{achlioptas2020referit3d} and ScanRefer~\cite{chen2020scanrefer} have built benchmarks and baselines for the 3D visual grounding task with language query annotations on ScanNet~\cite{dai2017scannet} dataset. Recent works~\cite{yuan2021instancerefer,achlioptas2020referit3d,chen2020scanrefer,he2021transrefer3d,roh2021languagerefer,yang2021sat} solve the 3D grounding tasks following a two-stage scheme. In the first stage, 3D object proposals are generated either with ground truth~\cite{achlioptas2020referit3d} or by a 3D object detector~\cite{chen2020scanrefer,jiang2020pointgroup}. In the second stage, the task is modeled as a matching problem where 3D visual features are fused with the text features of the query language description to predict the matching scores. Referit3D~\cite{achlioptas2020referit3d} uses GRU~\cite{chung2014empirical} to extract text features, and makes use of GNN~\cite{scarselli2008graph} to model the relationship between objects. With the widespread of transformers~\cite{vaswani2017attention}, recent works~\cite{yang2021sat,he2021transrefer3d,roh2021languagerefer} have explored transformers for feature extraction and feature fusion. Among them, InstanceRefer~\cite{yuan2021instancerefer} conducts multi-level contextual referring by considering attributes and local relations. LanguageRefer~\cite{roh2021languagerefer} replaces point cloud features with predicted labels of objects, transforming the multi-modal task into a language modeling problem. SAT~\cite{yang2021sat} has conducted joint training of 2d grounding and 3d grounding, improving the performance significantly by cross-modal knowledge transfer.
3DVG-Transformer~\cite{zhao20213dvg} proposes a coordinate-guided contextual aggregation module and a multiplex attention module to fully utilize the contextual clues in the 3D scene. 
FFL-3DOG~\cite{feng2021free} uses language and visual scene graphs to guide the feature matching between language and point cloud.
TransRefer3D~\cite{he2021transrefer3d} designs an entity-aware attention module and a relation-aware attention module to conduct fine-grained cross-modal feature matching.
Different from previous works, we start by studying the intrinsic problem of 3D visual data and build a view-robust multi-modal representation to solve the problem.

\noindent \textbf{Learning with Natural Language Supervision.}
In parallel, there are also a series of works~\cite{joulin2016learning,zhang2020contrastive,desai2021virtex,gomez2017self,radford2021learning} to learn the visual model guided by natural language features, with the target of learning transferrable models from language. Among them, VirTex~\cite{desai2021virtex} pre-trains the network using dense captions (e.g., COCO Captions~\cite{chen2015microsoft}) to learn visual representations and transfer them to downstream visual recognition tasks.
CLIP~\cite{radford2021learning} performs the natural language supervision on a sufficiently large dataset collected from the Internet and demonstrates its powerful zero-shot transfer capability. In this paper, we leverage fine-grained object labels to enhance object encoders to improve the alignment between object features and language queries.

\noindent\textbf{Multi-View Learning.} In 3D computer vision, recent works have focused on learning better representation by rendering objects from different views.
MV3D~\cite{chen2017multi} encodes the sparse 3D point cloud with a compact multi-view (e.g., bird's eye view and front view) representation for the 3D detection task~\cite{Geiger2012CVPR}.
MVCNN~\cite{su2015multi} completes the 3d object classification by rendering a large number of 2d pictures from 3d objects. These works project 3D point clouds onto different 2D planes to enhance visual features. Unlike them, we project the positional information of the 3D scene into a multi-view space, thereby learning a view-robust representation for better vision-language alignment.

\section{Method}

\begin{figure*}[t]
  \centering
  \includegraphics[width=0.99\textwidth]{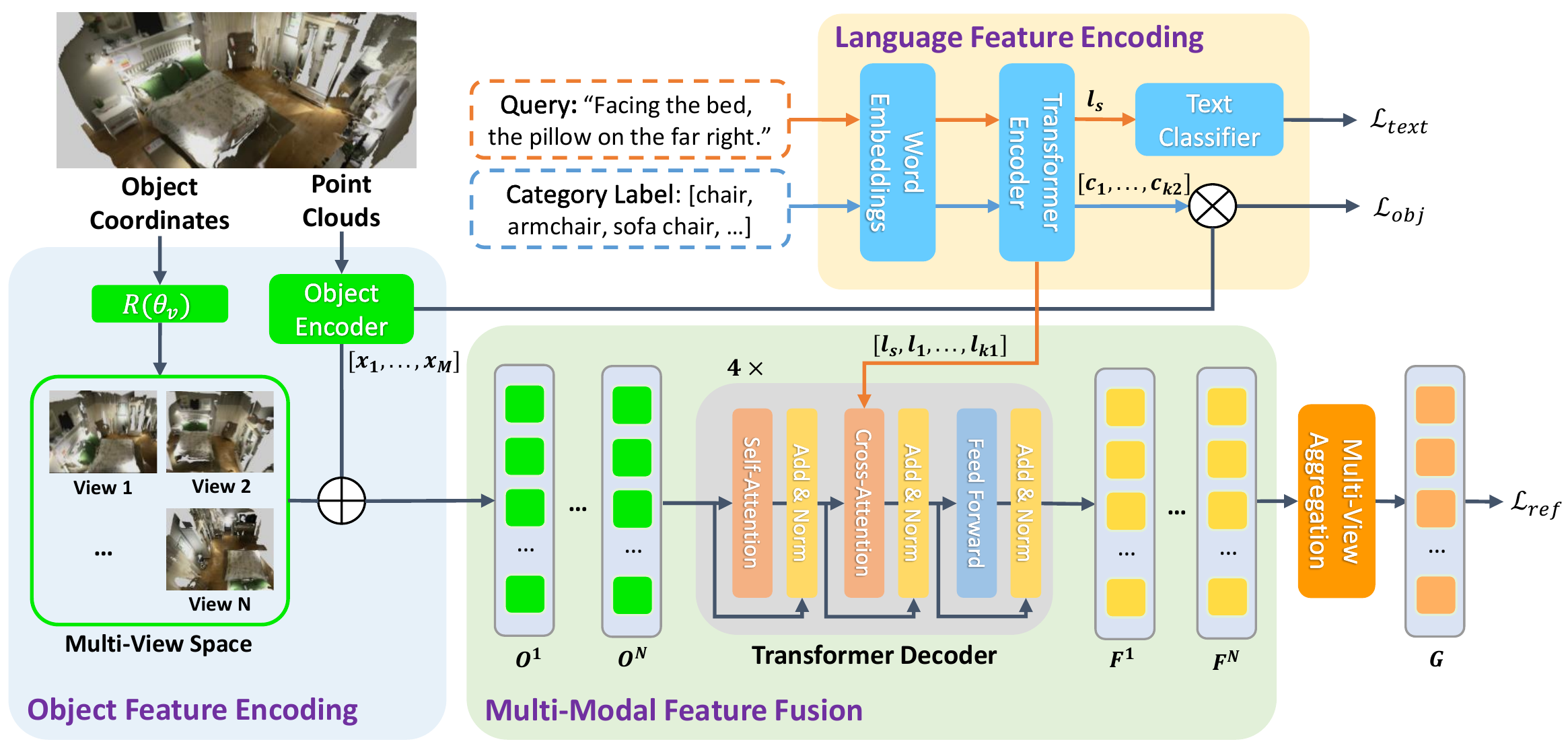}
  \vspace{-5pt}
  \caption{The framework of the proposed multi-view transformer (MVT) for 3D visual grounding. We project the 3D scene to a multi-view space by the equal angle rotation. For each view, object features and language features are fused by the transformer decoder~\cite{vaswani2017attention}. We finally aggregate information from multi-view space to form a view-robust representation and predict the grounding results.}
  \label{fig:framework}
  \vspace{-5pt}
\end{figure*}

In this section, we introduce the proposed Multi-View Transformer (MVT) for 3D visual grounding. 
Fig.~\ref{fig:framework} shows the whole framework, which contains object encoding, language encoding, multi-modal feature fusion, and multi-view aggregation modules. 
In this section, we first illustrate our multi-view 3D visual grounding framework and then provide details of each component.

\subsection{Multi-View 3D Visual Grounding}
\label{sec:view_indep_3DVG}

\begin{figure}[t]
  \centering
  \includegraphics[width=0.40\textwidth]{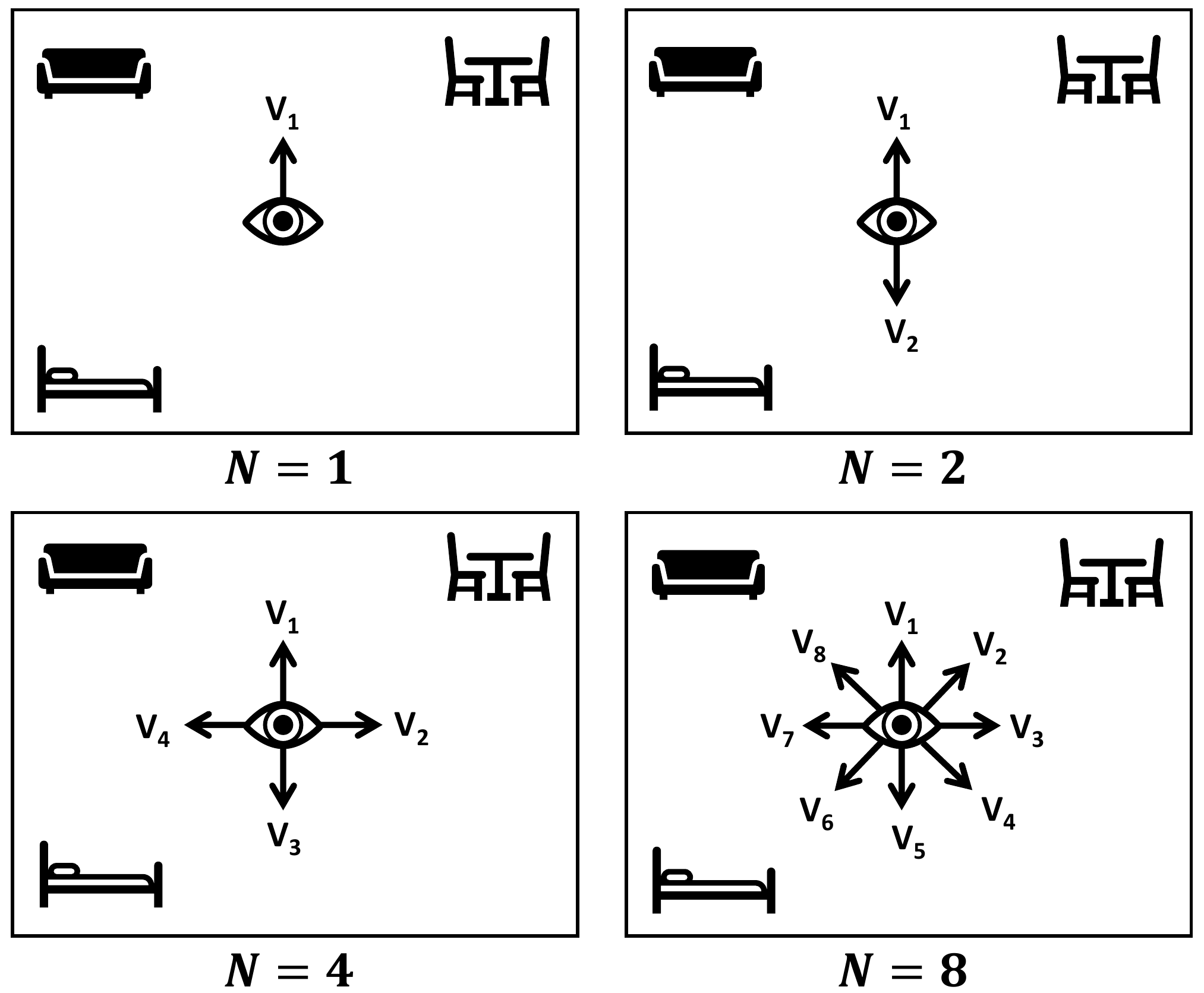}
  \vspace{-5pt}
  \caption{The illustration of multi-view generation. $N$ is the view number, $V_{1}$ is the initial view of scene $S$. We generate multiple views by the equal angle rotation, which divides the $2\pi$ view space equally into $N$ parts.}
  \label{fig:view}
  \vspace{-5pt}
\end{figure}

The key idea of our model is to learn a multi-modal representation independent from its specific starting view. Given the 3D scene $\rm{S}$ under an arbitrary view and a natural language query
$\rm{Q}$, we extend $\rm{S}$ to $N$ different views $\{\rm{S}^{1},\dots,\rm{S}^{N}\}$ through rotations, given as:
\begin{equation}
    \text{$\rm{S}$}^{j} = R(\theta^j_{v}) \times \rm{S},
\end{equation}
\begin{equation}
    R(\theta) = \begin{pmatrix}
      \cos{\theta} & -\sin{\theta} & 0 \\ 
      \sin{\theta} & \cos{\theta} & 0 \\
      0 & 0 & 1 
    \end{pmatrix}^T,
    \label{equ:rotate_matrix}
\end{equation}
\noindent where $R(\theta)$ is the rotation matrix, and $T$ is the transpose. $R(\theta) \times \rm{S}$ means rotating the 3D scene $\rm{S}$ clockwise by $\theta$ degrees along the $Z$-axis. $\theta^{j}_{v}$ is the view-rotation angle of the $j$-th view. We adopt the equal angle rotation, which is:
\begin{equation}
    \theta^{j}_{v} = \frac{2 \pi (j-1)}{N}, j \in \{1,\dots,N\},
    \label{equ:thetav}
\end{equation}
\noindent where $N$ is the number of total views, usually set to be $\{1, 2, 4, 8\}$. Fig.~\ref{fig:view} shows $\theta_{v}$ with different view settings. $\theta^{1}_{v}$ equals to $0$ hence $\rm{S}^{1}$ is the same as $\rm{S}$.
Therefore, the original 3D visual grounding task $\rm{VG}$ under certain view $\rm{S}$ is now extended to learn a more robust multi-view representation, which is modeled as:
\begin{equation}
    \text{$\rm{VG}$}(\text{$\rm{S}$},Q) = \text{$\rm{g}$}(f(\text{$\rm{S}$}^{1},Q),\dots,f(\text{$\rm{S}$}^{N},Q)),
    \label{equ:define}
\end{equation}
\noindent where $f(\cdot)$ is a shared network for feature extractions under different views. $f(\cdot)$ takes into account the position information of different views in the multi-view space.
$\rm{g}(\cdot)$ is an order-independent function for information aggregation. Therefore, Eq.(\ref{equ:define}) has the intriguing property, i.e., when we change the starting view from $\rm{S}$ to any other in $\{\rm{S}^{1}, \dots, S^{N}\}$, the final multi-view representation of $\rm{VG}(S,Q)$ keeps invariant. This property meets our goal of building a robust representation that is independent 
of its initial view by projecting $\rm{S}$ to the multi-view space of $\{\rm{S}^{1},\dots,\rm{S}^{N}\}$.


\subsection{Object Feature Encoding}

Following previous settings~\cite{achlioptas2020referit3d,chen2020scanrefer}, we assume to have access to $M$ candidate objects in the scene $\rm{S}$ under a specific view. Objects are either generated from the ground truth as in Referit3D~\cite{achlioptas2020referit3d} or predicted by a 3D object detector as in ScanRefer~\cite{chen2020scanrefer}.
3D object encoding extracts features from 3D point cloud data, which is the main computing bottleneck, and it is more costly to encode features for multiple views. Therefore, we decoupled the 3D object feature extraction process into two steps, i.e., computing \textit{point clouds features} and \textit{multi-view positional encoding}. This decoupling can significantly save the computational cost since different views can share the same point cloud features outputted from the first step. Finally, the object features can be calculated by simply combining both.

\noindent \textbf{Point Cloud Features.} We sample $1024$ points of point clouds for each object $i$ in the form of RGB-XYZ, denoted as $pc_{i} \in \mathbb{R}^{1024 \times 6}$. Then point cloud features $\{x_{1},\dots,x_{M}\}$ for all objects are extracted by the object encoder (e.g. PointNet++~\cite{qi2017pointnet++}) and a linear mapping layer, given as:
\begin{equation}
    x_{i} = \text{LN}(W_{x} \cdot \text{PointNet++}(pc_{i})),
    \label{equ:obj_feat}
\end{equation}
\noindent where $x_{i}$ is the point cloud feature of object $i$, $W_{x}$ is a projection matrix. $\text{LN}(\cdot)$ is layer normalization~\cite{ba2016layer}.

\noindent \textbf{Multi-View Positional Encoding.} The object coordinates are represented by box center coordinates $\{b_{1},\dots,b_{M}\}$, in the form of XYZ. 
We rotate them to generate the object coordinates under multiple views, formulated as:
\begin{equation}
    b^{j}_{i}= R(\theta^{j}_{v}) \times b_{i}, 
    \label{equ:rotate_box}
\end{equation}
\noindent where $b^{j}_{i}$ is the coordinate of object $i$ under view $j$. $R(\theta)$ and $\theta^{j}_{v}$ are defined in Eq. (\ref{equ:rotate_matrix}) and Eq. (\ref{equ:thetav}). Then we obtain the $d$-dimensional positional encoding $\rm{PE}$$(b^{j}_{i})$ for object $i$ under view $j$, given as:
\begin{equation}
    \text{$\rm{PE}$}(b^{j}_{i}) = \text{LN}(W_{b}[b^{j}_{i},r_{i}]),
\end{equation}
\noindent where $r_{i} \in \mathbb{R}^{1}$ is the box size of object $i$, $W_{b}$ is a projection matrix mapping the box information to high dimension embedding, $\text{LN}(\cdot)$ is layer normalization~\cite{ba2016layer}, and $[\cdot,\cdot]$ is the concatenate operation. 

\noindent \textbf{Object Feature Generation.} We obtain final object features $O^{j} = \{o^{j}_{1},\dots,o^{j}_{M}\}$ under view $j$ by adding the positional encoding $\rm{PE}$ with the point cloud features $x$, given as:
\begin{equation}
    o^{j}_{i} = x_{i} + \text{$\rm{PE}$}(b^{j}_{i}),
\end{equation}
\noindent where $x_{i}$ is shared by multiple views, providing visual information, e.g., categories, colors, etc. $\rm{PE}(\cdot)$ provides the size and position information for each view.

\subsection{Multi-Modal Feature Fusion} 

The multi-modal features are fused by the language feature $L$ and the object features $O$ under multiple views. Following~\cite{yang2021sat,he2021transrefer3d,roh2021languagerefer}, we adopt a BERT model~\cite{devlin2018bert} as the language encoder. Given the query expression $Q$ with $k_1$ words, we embed them into $d$-dimensional feature vectors $L$, given as:
\begin{equation}
    \{l_{s}, l_{1}, \dots, l_{k_1}\} = \rm{BERT}(Q),
    \label{equ:lang_feat}
\end{equation}
\noindent where $l_{s}$ is the sentence-level feature, $l_{i}$ is the feature of $i$-th word. We fine-tune the BERT during training. To enhance the BERT, a text classifier with two FC layers is applied on $l_{s}$, which predicts the object category described by $Q$, and supervised by a cross-entropy loss $\mathcal{L}_{text}$.

We adopt a standard transformer decoder~\cite{vaswani2017attention} for the multi-modal feature fusion, which is shared by different views. We fuse the multi-modal feature under view $j$ as:
\begin{equation}
    F^{j} = \rm{Decoder}(O^{j},L),
\end{equation}
\noindent where $F^{j}=\{f^{j}_{1},\dots,f^{j}_{M}\}$ is multi-modal feature of object $i$ under view $j$. As shown at the bottom of Fig.~\ref{fig:framework}, in each layer of the decoder, object features $O^{j}$ first pass through a self-attention module~\cite{vaswani2017attention}, where interactions between objects are modeled. Then object features $O^{j}$ and language features $L$ are fused through a cross-attention module~\cite{vaswani2017attention}.

\subsection{Multi-View Aggregation} 

After the multi-modal feature fusion, we aggregate the multi-view information from  $\{f_{i}^{1},\dots,f_{i}^{N}\}$. The aggregation function should be order-independent as in Eq. (\ref{equ:define}), here we choose the average aggregation by default, given as:
\begin{equation}
    g_{i} = \sum_{j=1}^{N} \frac{f^{j}_{i}}{N},
    \label{equ:agg}
\end{equation}
\noindent where $G=\{g_{1},\dots,g_{M}\}$ is the final multi-modal feature of each object. Finally we obtain the grounding score by applying two FC layers on $G$.

We aggregate the multi-view information after the multi-modal fusion by default and then make the final prediction. We can also aggregate the multi-view information at earlier stages (e.g., after object feature encoding). There are also other alternatives to the aggregation function (e.g. $\max$, $\mathrm{avg}+\max$). We explore the performance of different aggregation stages and different aggregation functions in Tab.~\ref{tab:view_agg}.

\subsection{Improving Object Encoder}

To enhance the object encoder, we introduce an auxiliary task to learn the classification of 3D objects like previous works~\cite{achlioptas2020referit3d,roh2021languagerefer}. However, these works mainly apply a few FC layers on the object encoder to classify objects, using only the visual data but neglecting any multi-modal information in this visual grounding task. We propose the auxiliary task of language-guided classification by leveraging object category labels as the text supervision.

As shown in the top of Fig.~\ref{fig:framework}, given the set of category label texts (e.g. ``armchair'', ``rocking chair'', and ``sofa chair''), we first extract text features from category labels following Eq.(\ref{equ:lang_feat}). The $d$-dimensional category text features are denoted as $C=\{c_{1},\dots,c_{k_2}\} \in \mathbb{R}^{k_2 \times d}$, where $c_{i}$ is the sentence-level feature of category $i$, and $k_2$ is the category number (e.g. $524$ in Nr3D~\cite{achlioptas2020referit3d} and $607$ in Sr3D~\cite{achlioptas2020referit3d}). Then classification logits $p_{i} \in \mathbb{R}^{k_2}$ of object $i$ are calculated by the inner product of category text features $C$ and the object feature $x_{i}$. We adopt a cross-entropy loss to supervise $p_{i}$, denotes as $\mathcal{L}_{obj}$.


\subsection{Overall Loss Functions}

We use a cross-entropy loss $\mathcal{L}_{ref}$ to supervise the final grounding score, and two auxiliary losses $\mathcal{L}_{text}$ and $\mathcal{L}_{obj}$ mentioned above to enhance the object encoder and language encoder. The total loss is defined as:
\begin{equation}
    \mathcal{L}_{total} = \mathcal{L}_{ref} + \alpha(\mathcal{L}_{text} + \mathcal{L}_{obj}),
\end{equation}
\noindent where $\alpha$ controls the ratio of each loss term. We set $\alpha=0.5$ by default.

\section{Experiments}

\begin{table*}[t]
\centering
\footnotesize
\begin{tabular}{c|c|lllll}
\multicolumn{1}{c|}{Arch.} & \multicolumn{1}{c|}{Extra} & \multicolumn{1}{c}{Overall} & \multicolumn{1}{c}{Easy} & \multicolumn{1}{c}{Hard} & \multicolumn{1}{c}{View-dep.} & \multicolumn{1}{c}{View-indep.} \\ \toprule
\multicolumn{6}{c}{\textbf{Nr3D}}                                                 \\ \hline
ReferIt3D~\cite{achlioptas2020referit3d} & None & 35.6\%$\pm$0.7\% & 43.6\%$\pm$0.8\% & 27.9\%$\pm$0.7\% & 32.5\%$\pm$0.7\% & 37.1\%$\pm$0.8\% \\
InstanceRefer~\cite{yuan2021instancerefer} & None  & 38.8\%$\pm$0.4\% & 46.0\%$\pm$0.5\% & 31.8\%$\pm$0.4\% & 34.5\%$\pm$0.6\% & 41.9\%$\pm$0.4\% \\
3DVG-Transformer~\cite{zhao20213dvg} & None & 40.8\% $\pm$ 0.2\% & 48.5\% $\pm$ 0.2\% & 34.8\% $\pm$ 0.4\% & 34.8\% $\pm$ 0.7\% & 43.7\% $\pm$ 0.5\% \\
FFL-3DOG~\cite{feng2021free} & None & 41.7\%     & 48.2\%     & 35.0\%       & 37.1\%      & 44.7\%        \\
TransRefer3D~\cite{he2021transrefer3d} & None & 42.1\%$\pm$0.2\% & 48.5\%$\pm$0.2\% & 36.0\%$\pm$0.4\% & 36.5\%$\pm$0.6\% & 44.9\%$\pm$0.3\% \\
LanguageRefer~\cite{roh2021languagerefer} & None  & 43.9\%     & 51.0\%       & 36.6\%     & 41.7\%      & 45.0\%         \\
SAT~\cite{yang2021sat} & 2D assist.  & 49.2\%$\pm$0.3\% & 56.3\%$\pm$0.5\% & 42.4\%$\pm$0.4\% & 46.9\%$\pm$0.3\% & 50.4\%$\pm$0.3\% \\ \hline
Ours & None  & \makecell[c]{\textbf{55.1\%$\pm$0.3\%} \\ \footnotesize \textcolor[RGB]{102,178,84}{(+11.2\%)}} & \makecell[c]{\textbf{61.3\%$\pm$0.4\%} \\ \footnotesize \textcolor[RGB]{102,178,84}{(+11.3\%)}} & \makecell[c]{\textbf{49.1\%$\pm$0.4\%} \\ \footnotesize \textcolor[RGB]{102,178,84}{(+12.5\%)}} & \makecell[c]{\textbf{54.3\%$\pm$0.5\%} \\ \footnotesize \textcolor[RGB]{102,178,84}{(+12.6\%)}} &  \makecell[c]{\textbf{55.4\%$\pm$0.3\%} \\ \footnotesize \textcolor[RGB]{102,178,84}{(+10.4\%)}} \\ \toprule
\multicolumn{6}{c}{\textbf{Nr3D w/ Sr3D}} \\ \hline
ReferIt3D~\cite{achlioptas2020referit3d} & None  & 37.2\%$\pm$0.3\% & 44.0\%$\pm$0.6\% & 30.6\%$\pm$0.3\% & 33.3\%$\pm$0.6\% & 39.1\%$\pm$0.2\% \\
non-SAT~\cite{yang2021sat} & None  & 43.9\%$\pm$0.3\% & -        & -        & -         & -           \\
TransRefer3D~\cite{he2021transrefer3d} & None  & 47.2\%$\pm$0.3\% & 55.4\%$\pm$0.5\% & 39.3\%$\pm$0.5\% & 40.3\%$\pm$0.4\% & 50.6\%$\pm$0.2\% \\
SAT~\cite{yang2021sat} & 2D assist.  & 53.9\%$\pm$0.2\% & 61.5\%$\pm$0.1\% & 46.7\%$\pm$0.3\% & 52.7\%$\pm$0.7\% & 54.5\%$\pm$0.3\% \\ \hline 
Ours & None  & \makecell[c]{\textbf{58.5\%$\pm$0.2\%} \\ \footnotesize \textcolor[RGB]{102,178,84}{(+11.3\%)}} & \makecell[c]{\textbf{65.6\%$\pm$0.2\%} \\ \footnotesize \textcolor[RGB]{102,178,84}{(+10.2\%)}} & \makecell[c]{\textbf{51.6\%$\pm$0.3\%} \\ \footnotesize \textcolor[RGB]{102,178,84}{(+12.3\%)}} & \makecell[c]{\textbf{56.6\%$\pm$0.3\%} \\ \footnotesize \textcolor[RGB]{102,178,84}{(+16.3\%)}} &  \makecell[c]{\textbf{59.4\%$\pm$0.2\%} \\ \footnotesize \textcolor[RGB]{102,178,84}{(+8.8\%)}} \\ \toprule
\multicolumn{6}{c}{\textbf{Nr3D w/ Sr3D+}}                                        \\ \hline
ReferIt3D~\cite{achlioptas2020referit3d} & None & 37.6\%$\pm$0.4\% & 45.4\%$\pm$0.6\% & 30.0\%$\pm$0.4\% & 33.1\%$\pm$0.5\% & 39.8\%$\pm$0.4\% \\
non-SAT~\cite{yang2021sat} & None & 45.9\%$\pm$0.2\% & -        & -        & -         & -           \\
TransRefer3D~\cite{he2021transrefer3d} & None & 48.0\%$\pm$0.2\% & 56.7\%$\pm$0.4\% & 39.6\%$\pm$0.2\% & 42.5\%$\pm$0.2\% & 50.7\%$\pm$0.4\% \\
SAT~\cite{yang2021sat} & 2D assist. & 56.5\%$\pm$0.1\% & 64.9\%$\pm$0.2\% & 48.4\%$\pm$0.1\% & 54.4\%$\pm$0.3\% & 57.6\%$\pm$0.1\%  \\ \hline
Ours & None  & \makecell[c]{\textbf{59.5\%$\pm$0.2\%} \\ \footnotesize \textcolor[RGB]{102,178,84}{(+11.5\%)}} & \makecell[c]{\textbf{67.4\%$\pm$0.4\%} \\ \footnotesize \textcolor[RGB]{102,178,84}{(+10.7\%)}} & \makecell[c]{\textbf{52.7\%$\pm$0.4\%} \\ \footnotesize \textcolor[RGB]{102,178,84}{(+13.1\%)}} & \makecell[c]{\textbf{59.1\%$\pm$0.5\%} \\ \footnotesize \textcolor[RGB]{102,178,84}{(+16.6\%)}} &  \makecell[c]{\textbf{60.3\%$\pm$0.1\%} \\ \footnotesize \textcolor[RGB]{102,178,84}{(+9.6\%)}} \\ \toprule
\end{tabular}
\vspace{-5pt}
\caption{Performance on Nr3D trained with or without Sr3D/Sr3D+. The improvement is calculated in comparison with the best competitor without extra assistance.}
\label{tab:Nr3d}
\vspace{-5pt}
\end{table*}

\begin{table*}[ht]
\footnotesize
\centering
\begin{tabular}{c|c|lllll}
\multicolumn{1}{c|}{Arch.} & \multicolumn{1}{c|}{Extra} & \multicolumn{1}{c}{Overall} & \multicolumn{1}{c}{Easy} & \multicolumn{1}{c}{Hard} & \multicolumn{1}{c}{View-dep.} & \multicolumn{1}{c}{View-indep.} \\ \bottomrule
ReferIt3D~\cite{achlioptas2020referit3d} & None & 40.8\%$\pm$0.2\% & 44.7\%$\pm$0.1\% & 31.5\%$\pm$0.4\% & 39.2\%$\pm$1.0\% & 40.8\%$\pm$0.1\% \\
InstanceRefer~\cite{yuan2021instancerefer} & None & 48.0\%$\pm$0.3\% & 51.1\%$\pm$0.2\% & 40.5\%$\pm$0.3\% & 45.4\%$\pm$0.9\% & 48.1\%$\pm$0.3\% \\
3DVG-Transformer~\cite{zhao20213dvg} & None & 51.4\% $\pm$ 0.1\% & 54.2\% $\pm$ 0.1\% & 44.9\% $\pm$ 0.5\% & 44.6\% $\pm$ 0.3\% & 51.7\% $\pm$ 0.1\% \\
LanguageRefer~\cite{roh2021languagerefer} & None & 56.0\% & 58.9\% & 49.3\% & 49.2\% & 56.3\% \\
TransRefer3D~\cite{he2021transrefer3d} & None & 57.4\%$\pm$0.2\% & 60.5\%$\pm$0.2\% & 50.2\%$\pm$0.2\% & 49.9\%$\pm$0.6\% & 57.7\%$\pm$0.2\% \\
SAT $\dag$~\cite{yang2021sat} & 2D assist. & 57.9\% & 61.2\% & 50.0\% & 49.2\% & 58.3\%  \\ \hline
Ours & None  & \makecell[c]{\textbf{64.5\%$\pm$0.1\%} \\ \footnotesize \textcolor[RGB]{102,178,84}{(+7.1\%)}} & \makecell[c]{\textbf{66.9\%$\pm$0.1\%} \\ \footnotesize \textcolor[RGB]{102,178,84}{(+6.4\%)}} & \makecell[c]{\textbf{58.8\%$\pm$0.1\%} \\ \footnotesize \textcolor[RGB]{102,178,84}{(+8.6\%)}} & \makecell[c]{\textbf{58.4\%$\pm$0.8\%} \\ \footnotesize \textcolor[RGB]{102,178,84}{(+8.5\%)}} &  \makecell[c]{\textbf{64.7\%$\pm$0.1\%} \\ \footnotesize \textcolor[RGB]{102,178,84}{(+7.0\%)}} \\ \toprule 
\end{tabular}
\vspace{-5pt}
\caption{Performance on Sr3D. The improvement is calculated in comparison with the best competitor without extra assistance.}
\label{tab:Sr3d}
\vspace{-5pt}
\end{table*}

\begin{table*}[ht]
\footnotesize
\centering
\begin{tabular}{c|cc|cc|cc}
\multicolumn{1}{c|}{\multirow{2}{*}{Method}} & \multicolumn{2}{c|}{Unique} & \multicolumn{2}{c|}{Multiple} & \multicolumn{2}{c}{Overall} \\ 
\multicolumn{1}{c|}{} & \multicolumn{1}{c}{Acc@0.25} & \multicolumn{1}{c|}{Acc@0.5} & Acc@0.25 & \multicolumn{1}{c|}{Acc@0.5} & Acc@0.25      & \multicolumn{1}{c}{Acc@0.5}     \\ \bottomrule 
\multicolumn{1}{c|}{ScanRefer~\cite{chen2020scanrefer}} & 65.00\% & 43.31\% & 30.63\% & 19.75\% & 37.30\% & 24.32\% \\ 
\multicolumn{1}{c|}{TGNN~\cite{huang2021text}} & 64.50\% & 53.01\% & 27.01\% & 21.88\% & 34.29\% & 27.92\% \\ 
\multicolumn{1}{c|}{TGNN~\cite{huang2021text} + BERT~\cite{devlin2018bert}} & 68.61\% & 56.80\% & 29.84\% & 23.18\% & 37.37\% & 29.70\% \\ 
\multicolumn{1}{c|}{IntanceRefer~\cite{yuan2021instancerefer}} & 77.45\% & \textbf{66.83\%} & 31.27\% & 24.77\% & 40.23\% & 32.93\% \\ \hline
\multicolumn{1}{c|}{Ours ($\text{view num = }1$)} & 74.36\% & 63.04\% & 29.65\% & 23.44\% & 38.33\% & 31.12\% \\ 
\multicolumn{1}{c|}{Ours ($\text{view num = }4$)} & \textbf{77.67\%} & 66.45\% & \textbf{31.92\%} & \textbf{25.26\%} & \textbf{40.80}\% & \textbf{33.26\%} \\ \toprule
\end{tabular}
\vspace{-5pt}
\caption{Performance on ScanRefer compared with previous works.}
\label{tab:ScanRefer}
\end{table*}

\begin{table*}[t]
\footnotesize
\centering
\begin{tabular}{c|cccc|ccccc}
     & \textbf{Decoder} & \textbf{Aug.} & \textbf{Lang-Sup.} & \textbf{Multi-View} &  Overall  & Easy     & Hard     & View-dep. & View-indep. \\ \bottomrule
 (a) &         &         &         &         & 36.9\% & 43.5\% & 30.5\% & 32.9\% & 38.8\% \\ \hline
 (b) & $\surd$ &         &         &         & 40.4\% & 47.5\% & 33.7\% & 38.4\% & 41.4\% \\
 (c) & $\surd$ & $\surd$ &         &         & 40.8\% & 48.4\% & 33.5\% & 35.2\%$\downarrow$ & 43.6\% \\ \hline
 (d) & $\surd$ & $\surd$ & $\surd$ &         & 46.2\% & 53.8\% & 38.9\% & 42.6\% & 48.0\% \\
 (e) & $\surd$ & $\surd$ &         & $\surd$ & 52.3\% & 59.1\% & 45.7\% & 50.3\%$\uparrow$ & 53.3\% \\ \hline
 (f) & $\surd$ &         & $\surd$ & $\surd$ & 53.3\% & 59.8\% & 47.0\% & 54.4\% & 52.7\% \\
 (g) & $\surd$ & $\surd$ & $\surd$ & $\surd$ & 55.1\% & 61.3\% & 49.1\% & 54.3\%$\approx$ & 55.4\% \\ \toprule
\end{tabular}
\vspace{-5pt}
\caption{Ablation studies of the MVT components on Nr3D.}
\label{tab:component_ablation}
\vspace{-5pt}
\end{table*}

\subsection{Datasets}

\noindent \textbf{Nr3D.} The Nr3D dataset~\cite{achlioptas2020referit3d} annotates the indoor 3D scene dataset ScanNet~\cite{dai2017scannet} with $45,503$ human utterances. There are a total of $707$ unique indoor scenes with target objects from $76$ fine-grained classes. Each scene contains no more than six distractors (objects in the same class as the target).
Two kinds of data splits are used in Referit3D. The ``easy'' and ``hard'' splits depend on whether the scene contains more than two distractors. The ``view-dep.'' and ``view-indep.'' splits depend on whether the referring expression is dependent on the speaker's view or not.

\noindent \textbf{Sr3D/Sr3D+.} The Sr3D dataset~\cite{achlioptas2020referit3d} contains $83,572$ utterances that are automatically generated based on a ``target-spatial relationship-anchor object'' template. The Sr3D+ dataset further expands Sr3D by adding samples without multiple distractors, resulting in a $114,532$ utterances.

\noindent \textbf{ScanRefer}. The ScanRef dataset~\cite{chen2020scanrefer} annotates $800$ indoor scenes in ScanNet~\cite{dai2017scannet} with $51,583$ human utterances, containing $36,665$, $9,508$, and $5,410$ samples in train/val/test sets, respectively. The data can be divided into ``Unique'' and ``Multiple'', depending on whether there are objects of the same class as the target in the scene.

\subsection{Experimental Setting}

\noindent \textbf{Evaluation metric.} For datasets in Referit3D~\cite{achlioptas2020referit3d} (e.g. Nr3D, Sr3D, and Sr3D+), the proposals are generated from the ground truth. The models are evaluated by the accuracy, which measures the percentage of successful matches between the predicted proposal and the ground truth proposal. For ScanRefer~\cite{chen2020scanrefer}, the proposals are generated from a pre-trained detector. We adopt PointGroup~\cite{jiang2020pointgroup} as the detector, following InstanceRefer~\cite{yuan2021instancerefer}. The Acc@$m$IoU is adopted as the evaluation metric, where $m\text{IoU} \in \{0.25, 0.5\}$. Acc@$m$IoU measures the fraction of language queries whose predicted box overlaps the ground truth box with 3D intersection over the union (IoU) higher than $m$.

\noindent \textbf{Implementation details.} We set the dimension $d$ in all transformer layers as $768$. We adopt the first three layers of BERT~\cite{devlin2018bert} as the text encoder and a four layers transformer decoder~\cite{vaswani2017attention} for multi-modal feature fusion, which makes the number of parameters comparable to SAT~\cite{yang2021sat}. The text encoder is initialized from the first three layers of $\text{BERT}_{\text{BASE}}$~\cite{vaswani2017attention}. The fusion decoder is trained from scratch. The number of views $N$ is set to $4$.
Model optimization is conducted using Adam~\cite{kingma2014adam} optimizer with a batch size of $24$. We set an initial learning rate of $0.0005$ for the model, and the learning rate of the transformer layer is further multiplied by $0.1$. We reduce the learning rate by a multiplicative factor of $0.65$ every $10$ epochs after $40$ epochs for a total of $100$ epochs.

\subsection{3D Visual Grounding Results}
\label{sec:main_result}

\noindent \textbf{Nr3D.} Tab.~\ref{tab:Nr3d} reports the grounding accuracy on Nr3D~\cite{achlioptas2020referit3d} dataset. Our MVT greatly surpasses all previous methods. Compared with the best competitor~\cite{roh2021languagerefer} using the same training setting, MVT outperforms the state-of-the-art method by $+11.2\%$, from $43.9\%$ to $55.1\%$. Even when compared with SAT~\cite{yang2021sat}, which utilizes extra 2D semantics assisted its training, our MVT still outperforms it by $5.9\%$ absolute value.
Specifically, the high accuracy on the ``Easy'' split ($11.3\%$ better than LanguageRefer~\cite{roh2021languagerefer}) reflects the better visual-language alignment of our MVT. At the same time, the high accuracy on the ``View-dep'' split ($12.6\%$ better than LanguageRefer~\cite{roh2021languagerefer}) reflects the effectiveness and robustness of the multi-view representation learned by our MVT. Tab.~\ref{tab:Nr3d} also reports the performance of training on Nr3D together with Sr3D and Sr3D+ datasets~\cite{achlioptas2020referit3d}. Both synthetic datasets expand the scale of the training set, therefore further improving the performance on Nr3D. As the results show, our MVT on ``Nr3D w/ Sr3D'' and ``Nr3D w/ Sr3D+'' achieves the overall grounding accuracy of $58.5\%$ and $59.5\%$ respectively, which significantly surpasses the state-of-the-art method~\cite{he2021transrefer3d} by $11.3\%$ and $11.5\%$.

\noindent \textbf{Sr3D.} Tab.~\ref{tab:Sr3d} shows the grounding accuracy on Sr3D~\cite{achlioptas2020referit3d}. The MVT achieves great improvements and is significantly better than state-of-the-art methods~\cite{he2021transrefer3d,yang2021sat}. Though our approach does not use any additional 2D data, ours is still much better than SAT~\cite{yang2021sat} which is with 2D semantics assisting its training. Compared with TranRefer3D~\cite{he2021transrefer3d} and SAT~\cite{yang2021sat}, our MVT exceeds them $+7.1\%$ and $+6.6\%$ respectively. 

\noindent \textbf{ScanRefer.} Tab.~\ref{tab:ScanRefer} shows the performance on ScanRefer~\cite{chen2020scanrefer}. 
We follow the same settings with InstanceRefer~\cite{yuan2021instancerefer}. Since the detector has completed the 3D object classification and feature extraction steps, we only test the performance of the multi-view representation.
As the results show, the overall Acc@0.25 and Acc@0.5 of our MVT with the view number of $1$ is $38.33\%$ and $31.12\%$, respectively. After increasing the view number to be $4$, the performance is improved to $40.80\%$ and $33.26\%$ respectively.

\noindent \textbf{Inference Time.} On a TITAN X (Pascal), for one data pair of a 3D scene and a language description, the average inference time of MVT with view number $N=4$ is $264$ms, and the inference time with $N=1$ is $261$ms. The increase in inference time by our multi-view modeling is slight.
\subsection{Ablation studies}

\subsubsection{Effectiveness of each Component}
\label{sec:ablation}
\vspace{-5pt}

To investigate the effectiveness of each component in MVT, we conduct detailed ablation studies on Nr3D~\cite{achlioptas2020referit3d}, shown in Tab.~\ref{tab:component_ablation}. 
The baseline model in row (a) simply passes object features through several linear layers and then fuses object features with language features through multiplication directly. Thereby the baseline model does not model the relationship between objects. After we introduce the transformer decoder~\cite{vaswani2017attention} to model object relationship and vision-language fusion in row (b), the performance is improved from $36.9\%$ to $40.4\%$. 
We add random rotation augmentation $R(\theta_{aug})$ to the input scene in row (c), where $\theta_{aug}$ is randomly sampled from $\theta_{v}$ in Eq.(\ref{equ:thetav}) with $N=4$.
The improvement in the ``Easy'' split shows that the augmentation improves the performance of the object encoder. However, the rotation augmentation cannot eliminate the dependence on the specific views, and the performance in the ``View-dep'' split degraded from $38.4\%$ to $35.2\%$.
Row (d) and (e) bear out the effectiveness of the proposed language-guided supervision and multi-view modeling, respectively. They improve the performance to $46.2\%$ and $52.3\%$ respectively. The improvement of language-guided supervision comes from better vision-language alignments. The giant jump of the ``View-dep.'' split from $35.2\%$ to $49.1\%$ shows that our proposed multi-view space modeling effectively learns a view-robust representation, eliminating the dependence on specific views. Adopting both language-guided supervision and multi-view encoding in (g) can improve the overall accuracy to $55.1\%$, surpassing all previous methods. 
We also compare the performance w/ and w/o rotation augmentation after adding the multi-view modeling. Comparing the performance in row (f) and row (g), we find that after adding the multi-view modeling, the rotation augmentation can enhance the object encoder and cause no degradation in the ``View-dep.'' split.

\begin{figure*}[t]
  \centering
  \includegraphics[width=0.99\textwidth]{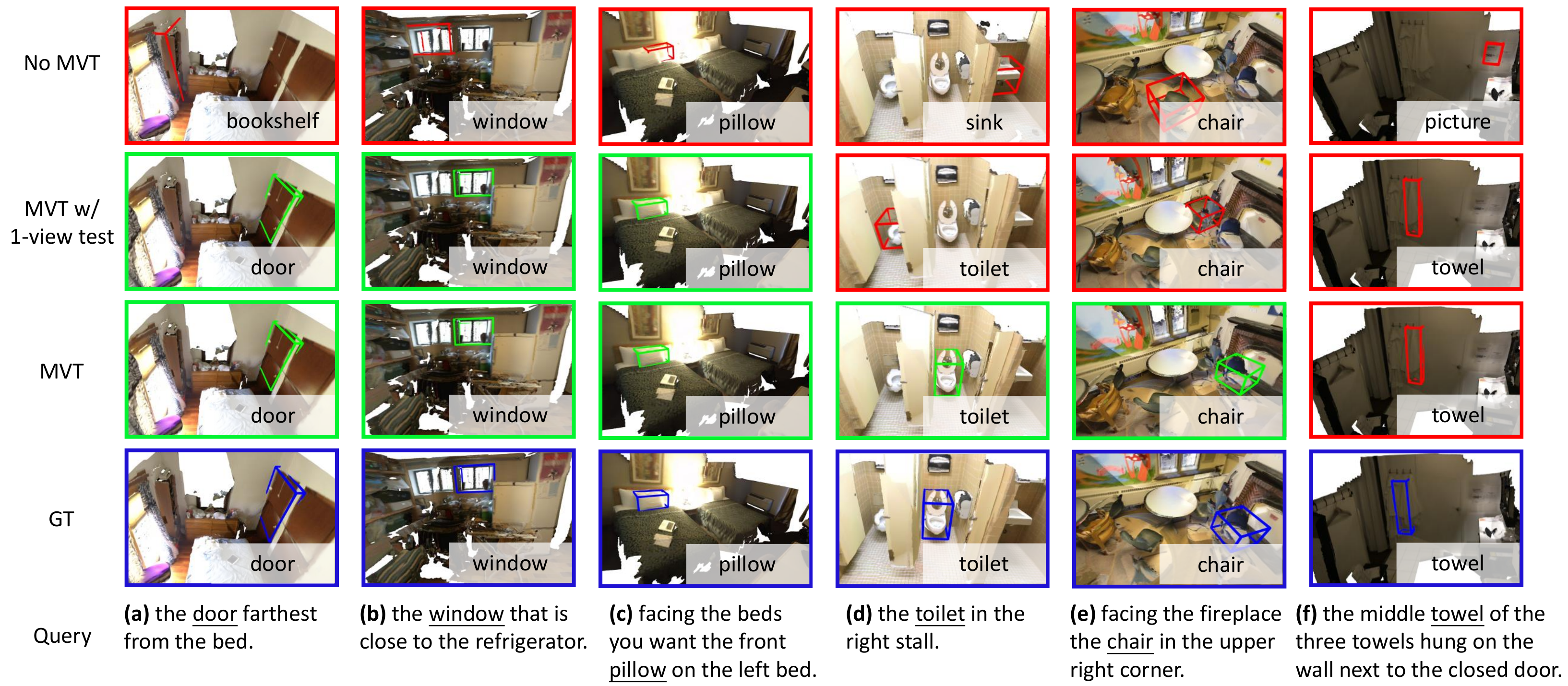}
  \vspace{-5pt}
  \caption{The visualization results. The green/red/blue colors indicate the correct/incorrect/ground truth boxes. Best viewed in color.}
  \label{fig:vis}
  \vspace{-5pt}
\end{figure*}

\begin{table}[ht]
\small
\centering
\resizebox{8.00cm}{!}{
\begin{tabular}{cc|ccccc}
\multicolumn{2}{c|}{View Number}     & \multicolumn{1}{c}{\multirow{2}{*}{Overall}} & \multicolumn{1}{c}{\multirow{2}{*}{Easy}} & \multicolumn{1}{c}{\multirow{2}{*}{Hard}} & \multicolumn{1}{c}{\multirow{2}{*}{View-dep.}} & \multicolumn{1}{c}{\multirow{2}{*}{View-indep.}}\\ 
\multicolumn{1}{c}{Train}  & \multicolumn{1}{c|}{Test}   & \multicolumn{1}{c}{} & \multicolumn{1}{c}{} & \multicolumn{1}{c}{} & \multicolumn{1}{c}{} & \multicolumn{1}{c}{} \\ \bottomrule 
\multicolumn{1}{c|}{4}      & 1      & 51.6\% & 58.4\% & 45.0\% & 50.7\% & 52.0\% \\
\multicolumn{1}{c|}{4}      & 2      & 54.3\% & 60.6\% & 48.3\% & 53.7\% & 54.7\% \\
\multicolumn{1}{c|}{4}      & 4      & 55.1\% & 61.3\% & 49.1\% & 54.3\% & 55.4\% \\
\multicolumn{1}{c|}{4}      & 8      & 54.8\% & 61.1\% & 48.8\% & 54.2\% & 55.1\% \\ \hline
\multicolumn{1}{c|}{1}      & 1      & 46.2\% & 53.8\% & 38.9\% & 42.6\% & 48.0\% \\
\multicolumn{1}{c|}{2}      & 2      & 51.9\% & 60.2\% & 43.9\% & 50.7\% & 52.5\% \\
\multicolumn{1}{c|}{8}      & 8      & 53.4\% & 60.9\% & 46.2\% & 53.5\% & 53.4\% \\ \toprule
\end{tabular}
}
\vspace{-5pt}
\caption{Ablation of view numbers on Nr3D.}
\vspace{-13pt}
\label{tab:view_number}
\end{table}

\subsubsection{Effectiveness of Multi-View Modeling.}

\noindent \textbf{Number of views.} Tab.~\ref{tab:view_number} shows the performance with different numbers of view on Nr3D~\cite{achlioptas2020referit3d}. We first compare the performance of $4$-view training but test in different numbers of views, which shows that the improvement of MVT comes from two parts. Firstly, when training with $4$ views but testing with $1$ view, the overall accuracy $51.6\%$ is much higher than training in 1 view only, i.e., $46.2\%$, and the ``View-dep.'' accuracy also increases by $8.1\%$. This result shows that multi-view modeling can learn an effective representation that can benefit different views. Secondly, when increasing the view number during testing, the grounding accuracy can still be improved. This observation tells that the information aggregation from the multi-view space can be helpful. We also compare the results of training with other view numbers. When training with only $2$ views, the model's performance can achieve $51.9\%$. 
When two views are relatively close, the position information provided by them for visual grounding is similar and redundant, thus we found $4$-view can already provide a robust multi-view representation. When training with more views  (e.g., $N=8$ views), redundant views cannot improve performance continually and may cause a little performance degradation due to low training efficiency.

\begin{table}[ht]
\small
\centering
\resizebox{8.3cm}{!}{
\begin{tabular}{c|ccccc}
Aggregation & Overall & Easy & Hard & View-dep. & View-indep. \\ \bottomrule
After $\rm{PE}$ & 42.5\% & 49.5\% & 35.8\% & 37.4\% & 45.0\% \\ 
After $+x$ & 47.4\% & 54.2\% & 41.0\% & 44.3\% & 49.0\% \\ 
After $*L$ & 55.1\% & 61.3\% & 49.1\% & 54.3\% & 55.4\% \\ \bottomrule
Avg &  55.1\% & 61.3\% & 49.1\% & 54.3\% & 55.4\%  \\ 
Max & 52.7\% & 60.0\% & 45.7\% & 51.7\% & 53.2\% \\ 
Avg + Max & 55.2\% & 62.1\% & 48.5\% & 54.4\% & 55.6\% \\ \toprule
\end{tabular}
}
\vspace{-5pt}
\caption{Ablation of multi-view aggregation on Nr3D.}
\vspace{-5pt}
\label{tab:view_agg}
\end{table}

\noindent \textbf{Multi-view aggregation.} Tab.~\ref{tab:view_agg} shows performances of different multi-view aggregation settings. We first adopt average aggregation and test different aggregation stages. The positional encoding is just a linear mapping with LayerNorm~\cite{ba2016layer}. If we aggregate after the positional encoding generation ($\rm{PE}$), it is equivalent to averaging the coordinates of multiple views. It performs badly since the position information is destroyed. To aggregate after object feature generation ($+x$), we first pass the object features into a 2-layers transformer encoder~\cite{vaswani2017attention} and then fuse object features from different views. It keeps the position information but still can not achieve the best performance. Aggregation after multi-modal feature fusion ($*L$) achieves the best performance. We also test different aggregation functions. ``Avg'' is more effective than ``Max'', and ``Avg + Max'' achieves similar performance as ``Avg''.

\noindent \textbf{Visualization.} We show the visualization results in Fig.~\ref{fig:vis}. ``No MVT'' is the baseline model same as in Tab.~\ref{tab:component_ablation} (b), ``MVT w/ 1-view test'' is the MVT model but test under 1-view, ``MVT'' is our proposed model. From (a, d, f), we can see that the baseline model's object recognition ability is relatively poor, and it is easy to predict objects of the wrong categories. By learning a view-robust representation, MVT can effectively learn the position correspondence between 3D scenes and language queries, thus predicting correctly in (b, c). We can see from (d, e) that information aggregation from the multi-view space can achieve better performance. We also show the failure cases of MVT in (f). When faced with complicated language queries and spatial relationships, MVT may also predict objects of the wrong categories or in the wrong positions.

\section{Conclusion}
In this paper, we propose a Multi-View Transformer (MVT) for 3D visual grounding. In MVT, the given 3D scene is projected to a multi-view space, in which the position information of the 3D scene under different views are modeled simultaneously. Based on it, MVT learns a more robust multi-modal representation for 3D visual grounding, which eliminates the dependence on specific views. Extensive experiments show that our MVT outperforms all state-of-the-art methods by a large margin. Detailed ablation studies show the effectiveness of all components in our model.

\clearpage
{\small
\bibliographystyle{ieee_fullname}

}

\end{document}